\PassOptionsToPackage{table}{xcolor}
\documentclass{article}

\usepackage{arxiv}
\usepackage[utf8]{inputenc}
\usepackage[T1]{fontenc}

\usepackage{xcolor}
\definecolor{linkblue}{rgb}{0.21,0.49,0.74}
\usepackage[colorlinks,allcolors=linkblue]{hyperref}
\usepackage{url}
\usepackage{graphicx,booktabs,amsmath,amssymb,amsfonts,multirow,adjustbox,pifont,xspace,microtype}
\usepackage[font=small,labelfont=bf,labelsep=period]{caption}
\usepackage[numbers,sort&compress]{natbib}
\usepackage[capitalize]{cleveref}
\crefname{figure}{Fig.}{Figs.}
\crefname{table}{Tab.}{Tabs.}
\crefname{section}{Sec.}{Secs.}
\crefname{appendix}{App.}{Apps.}

\definecolor{oursbg}{HTML}{E6F4F1}
\definecolor{oursc}{HTML}{1E7D71}
\newcommand{\ours}{STILL\xspace}
\newcommand{\std}[1]{{\scriptsize\textcolor{black!55}{$\pm$#1}}}
\newcommand{\best}[1]{\textbf{#1}}
\newcommand{\second}[1]{\underline{#1}}
\newcommand{\gain}[1]{{\footnotesize\textcolor{oursc}{#1}}}
\newcommand{\dlt}[1]{{\scriptsize\textcolor{black!55}{(#1)}}}
\newcommand{\cmark}{\textcolor[HTML]{2CA25F}{\ding{51}}}
\newcommand{\xmark}{\textcolor[HTML]{E34A33}{\ding{55}}}
\newcommand{\dmark}{\textcolor{black!40}{$\circ$}}
\DeclareMathOperator*{\argmax}{arg\,max}

\title{Is It Still There? Place-Registered Verification\\Makes VLMs Reliable Scene-Change Detectors}

\usepackage{authblk}

\author{Jinjie Xie}
\author{Baihua Li}
\author{Qinggang Meng\thanks{Corresponding author.}}
\affil{Department of Computer Science, Loughborough University, United Kingdom\\
\texttt{\{j.xie6, b.li, q.meng\}@lboro.ac.uk}}
\date{}

\renewcommand{\shorttitle}{Is It Still There? Place-Registered Verification for Scene-Change Detection}
\hypersetup{
  pdftitle={Is It Still There? Place-Registered Verification Makes VLMs Reliable Scene-Change Detectors},
  pdfauthor={Jinjie Xie, Baihua Li, Qinggang Meng},
  pdfkeywords={scene change detection, vision-language models, place-registered verification},
}

\begin{document}
\maketitle

\begin{abstract}
A robot that returns to a space must tell which objects were added, removed or moved since its last visit, while its trajectory, viewpoints and lighting differ between the two visits.
We find that multi-image vision-language models (VLMs) fail at this task mainly because they cannot find the place where an object was.
A VLM recognises what each frame shows but cannot check whether an object is still at its place in the other capture.
It therefore reports objects that only one capture shows and misses most removed and moved objects.
We propose \ours, which reformulates change detection as object verification at a registered place.
3D geometry registers the place of each change hypothesis in the other capture, and one VLM answers, from a scene view and a local place view, whether the object is still at that place, trained with supervision of the same form.
Under the official SceneDiff protocol, \ours raises the main metric obj/sc AP from 24.0 to 37.8, and from 11.6 to 20.6 on the egocentric kitchen subset without any kitchen training data.
It improves F1 by more than 21 points over a VLM fine-tuned on the same annotations and adds only about 11\% to the runtime.
Code is available at \url{https://github.com/Oran9er/STILL}.
\end{abstract}


\begin{figure}[t]
\centering
\includegraphics[width=\linewidth]{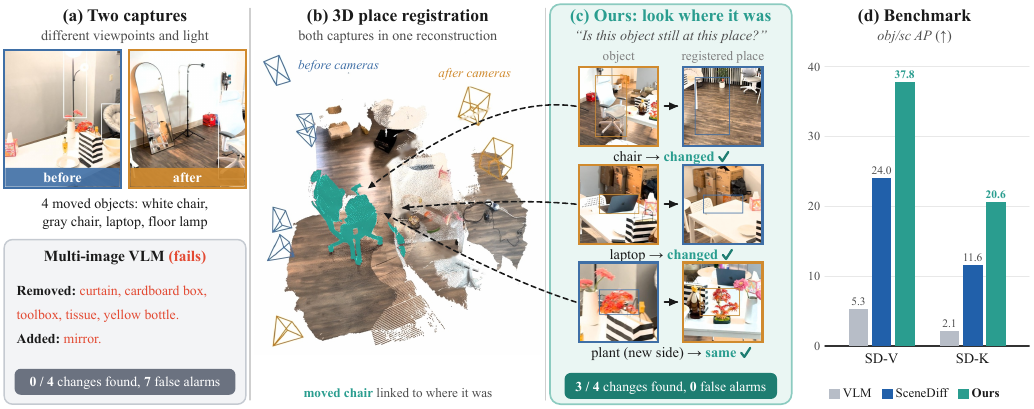}
\caption{\textbf{Change is a question of place.}
(a) Two captures of a room in which four objects were moved, under a new viewpoint and new lighting. A multi-image VLM fine-tuned on the same data reports objects that only one capture shows and finds none of the four.
(b) \ours registers each change hypothesis in a joint 3D reconstruction of both captures, which links the object to the place where it was.
(c) At that place, one VLM answers whether the object is still there. A moved chair leaves its place empty, while a plant seen from a new side is still in place.
(d) obj/sc AP under the official SceneDiff protocol for the VLM baseline of~\cite{wu2026scenediff} (Qwen2.5-VL + SAM3), reproduced SceneDiff and \ours.}
\label{fig:teaser}
\end{figure}

\section{Introduction}
\label{sec:intro}

A robot that operates in a home or an office over long periods must determine, each time it returns to a room, which objects have been added, removed or moved since its last visit.
Its memory of the environment, and the tidying and object search that rely on it, depend on this answer~\cite{schmid2024khronos,liu2024ok}.
The difficulty is that two visits never record the room in the same way, with a different route, different framing and different light.
Most visual differences between the two captures therefore involve no change at all, while a removed object leaves only an empty patch of tabletop.
SceneDiff~\cite{wu2026scenediff} formulates this task as multi-view object-level change detection.
Given two videos of the same scene recorded at two times along different trajectories, a model must find the added, removed and moved objects and localise them in every frame.

Multi-image vision-language models (VLMs) are becoming a general interface through which embodied agents understand scenes and answer spatial questions~\cite{yang2025thinking,bai2025qwen25vltechnicalreport,bai2025qwen3}.
Giving both videos to a VLM and asking which objects changed is the most natural way to approach the task, and we tested it systematically on SceneDiff.
Zero-shot Qwen3-VL~\cite{bai2025qwen3}, Qwen3.5 and InternVL3.5~\cite{wang2025internvl3} reach an F1 of only 9.2 to 14.3, and they report changes on 83\% to 93\% of control pairs built by splitting a single video, where nothing changed.
Fine-tuning on the same annotations lifts F1 to 39.2, but recall stays highly uneven, with 57\% on added objects against 36\% on removed and 34\% on moved objects.
The fine-tuned model still reports changes on 70\% of first-vs-second-half controls, whose two halves show different parts of the scene.

These errors share one cause.
To decide whether an object was removed, one has to look at the place where it was in the other capture.
If that place is empty, the object was removed.
If the same object is still there, seen from a new angle, nothing changed.
A holistic VLM is good at naming what an image contains, yet it cannot tell where an object should appear in the other video.
Curtains and mirrors that only one capture shows are therefore reported as added or removed, while the chair that actually moved is missed (\cref{fig:teaser}a).
A VLM judges whether an object is present, whereas change detection asks what is now at the place where the object was.

A direct idea is to train the VLM to find this place itself.
We trained a series of localisation models for this purpose, from per-frame localisation and zoomed refinement to reinforcement learning with a verification reward.
They recover about half of the places of unchanged objects and at most a quarter of those of changed objects, and SIFT matching~\cite{lowe2004distinctive} shows the same pattern.
All of them establish correspondence from the appearance of the object, and at a changed object this appearance is gone.
3D geometry does not have this limitation.
Feed-forward 3D reconstruction~\cite{wang2024dust3r,wang2025vggt,wang2026pi} places both captures in one coordinate frame, so the place where an object was follows from the static walls, tabletops and furniture around it, whether or not the object is still there.

These two observations lead to \ours, which reformulates change detection as object verification at a registered place.
Each change hypothesis becomes one question, whether the object is still at its place in the other capture, and this question determines what geometry computes, what the VLM sees and how the VLM is trained.
Hypotheses can come from any high-recall proposer, by default the 3D change detector of SceneDiff~\cite{wu2026scenediff}, and $\pi^3$~\cite{wang2026pi} lifts each one to 3D and projects it into the other capture, which gives its registered place.
One VLM then answers two local questions per hypothesis.
A scene view with all frames and a place view that shows the object next to its registered place both ask whether the object is still at its place, and the two answers and the geometric cost form the final score.
To train this verifier, we rewrite the holistic change annotations into per-object place questions, add an answer for places not in view, and use first-vs-second-half controls as negatives, with synthetic two-view pairs~\cite{sachdeva2023change} for two-view input.

Under the official SceneDiff protocol, \ours raises the main metric obj/sc AP from 24.0 to 37.8.
On the egocentric kitchen subset SD-K it raises obj/sc AP from 11.6 to 20.6 without any kitchen training data, relative gains of 57\% and 78\%.
Its F1 is 22 to 26 points higher than that of a VLM fine-tuned on the same annotations, and on the two-view benchmark RC-3D~\cite{sachdeva2023change} AP50 rises from 61.0 to 64.3.

Our main contribution is this reformulation, and the gains come from it.
The same untrained Qwen3.5 cuts its false alarms on first-vs-second-half controls from 93\% to 12\% once it is asked at the registered place.
The same annotations used for holistic fine-tuning, even with all per-object questions as auxiliary tasks, reach 40.5 F1, against 61.1 for \ours.
The formulation also works with a generic proposer, since with Grounding DINO~\cite{liu2024grounding} boxes the same verifier reaches 43.0 AP against 25.6 for SceneDiff, and a Qwen3-VL verifier performs on par.
We further show why the place must come from geometry, since appearance-based registration, learned or hand-crafted, fails exactly at changed objects.

\section{Related Work}
\label{sec:related}

\paragraph{Scene change detection.}
Early change detection compared pixels between image pairs taken from similar viewpoints, as in the street-view datasets VL-CMU-CD~\cite{alcantarilla2018street} and PSCD~\cite{sakurada2020weakly} and the indoor simulation ChangeSim~\cite{park2021changesim}.
CYWS~\cite{sachdeva2023changewacv} extended the problem to two images from different viewpoints and regressed boxes around changed objects.
CYWS-3D~\cite{sachdeva2023change} warped features between the views with estimated depth and released the real two-view benchmark RC-3D.
Robotic systems with depth sensors keep objects in a 3D map and compare their presence and pose on a revisit~\cite{schmid2024khronos,qian2022pocd}.
Rendering-based methods fit a radiance field or 3D Gaussians to the first capture and compare its renderings with the second capture~\cite{lu20253dgs,galappaththige2025multi}, which requires dense captures.
SceneDiff~\cite{wu2026scenediff} addresses two videos recorded along different trajectories, reconstructs them jointly with $\pi^3$~\cite{wang2026pi}, compares geometry and DINOv3 features on SAM regions without training, and released a multi-view benchmark with per-frame instance annotations.
\ours turns change regions, from SceneDiff or a generic detector, into verification questions at their registered places, which moves the decision from feature differences to a judgement of whether the object is still there.

\paragraph{Multi-image VLMs and change understanding.}
Change captioning asks a model to describe the differences between two images in language, with Spot-the-Diff~\cite{jhamtani2018learning} and CLEVR-Change~\cite{park2019robust} as representative benchmarks built on fixed cameras or controlled viewpoint shifts.
For multi-image VLMs, MuirBench~\cite{wang2025muirbench} and BLINK~\cite{fu2024blink} show a wide gap to humans on cross-image correspondence and multi-view reasoning.
Eyes Wide Shut~\cite{tong2024eyes} traces part of these failures to the visual encoder, and VSI-Bench~\cite{yang2025thinking} finds accurate local spatial relations but weak global layout in video.
In SceneDiff, the VLM baseline of Qwen2.5-VL with SAM3 correctly names only 113 of 369 changed objects.
We trace this failure to one step, finding where the object was in the other capture.

\paragraph{Geometry for VLMs.}
One line of work brings 3D information into the training data or the input of VLMs.
SpatialVLM~\cite{chen2024spatialvlm} generates large-scale spatial question answering from 3D reconstructions, 3D-LLM~\cite{hong20233d} feeds point clouds and multi-view features to the model, and Set-of-Mark~\cite{yang2023set} overlays region indices so that the model can refer to regions.
In \ours, geometry decides where to ask, and the model only sees ordinary frames and crops of one object and its registered place.
Any multi-image VLM can therefore serve as the verifier with standard supervised fine-tuning.

\paragraph{Feed-forward 3D reconstruction.}
DUSt3R~\cite{wang2024dust3r}, MASt3R~\cite{leroy2024grounding}, VGGT~\cite{wang2025vggt} and $\pi^3$~\cite{wang2026pi} predict per-pixel 3D points and camera poses directly from uncalibrated images and remain stable under small overlap and large viewpoint changes.
SceneDiff and \ours build on this ability to place two independent captures in one coordinate frame, and \ours further uses the reconstructed depth in a visibility test to find the frame in which the place of an object is actually visible.

\section{Why VLMs Miss Changes}
\label{sec:diagnosis}

We examine multi-image VLMs on the SD-V test split of SceneDiff, which contains 150 video pairs.
Each test item takes 5 frames from each of the two captures, and the model must list every changed object with its change type, frame index and box.
A predicted box with IoU of at least 0.3 with a ground-truth object box finds that object.
We also build two kinds of control pairs from a single video, in which nothing changed.
Interleaved controls split the frames into two interleaved groups with similar viewpoints.
First-vs-second-half controls split the video into halves that face different parts of the scene, so many objects are visible in only one group, as in real test items.
Any reported change on a control is a false alarm.

\begin{figure}[t]
\centering
\includegraphics[width=0.66\linewidth]{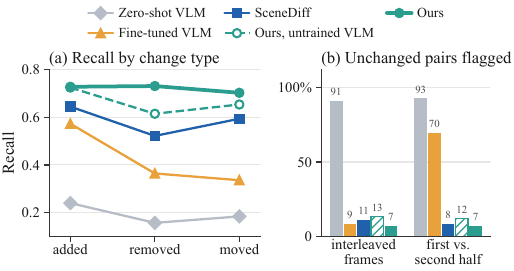}
\caption{\textbf{VLMs see what is present and miss what is gone} (SD-V test split, all VLMs are Qwen3.5-9B).
(a) Recall by change type at precision $\geq$0.34. The fine-tuned holistic VLM misses removed and moved objects, which the untrained VLM asked at the registered place (dashed) recovers.
(b) Share of no-change controls, cut from one video into interleaved frames or into two halves, with a reported change.}
\label{fig:diagnosis}
\end{figure}

Zero-shot VLMs report changes on almost every pair (\cref{fig:diagnosis}).
Qwen3-VL, Qwen3.5 and InternVL3.5 raise false alarms on 83\% to 92\% of interleaved controls, and on first-vs-second-half controls Qwen3.5 flags 268 of 289 pairs with on average 4 changes per pair that do not exist.

Fine-tuning Qwen3.5 on 40 video pairs of the SceneDiff dev split teaches it to stay silent on interleaved controls, where the false alarm rate drops to 9\%, and lifts F1 to 39.2.
Recall reaches 57\% on added objects but only 36\% on removed and 34\% on moved objects.
An added object is directly visible in the later capture, while a removed object can only be confirmed by seeing that its place is now empty, and a moved object needs both an empty old place and a newly occupied one.
The fine-tuned model also still reports changes on 70\% of first-vs-second-half controls, mostly objects that appear in only one group of frames, which is exactly the error expected when the place is unknown.

To test whether the place is the missing piece, we keep the VLM and change only how it is asked.
We first find the place of each change hypothesis in the other capture with 3D geometry and then ask an untrained Qwen3.5 whether the object is still at that place.
At the precision of SceneDiff, this zero-shot verifier recovers 62\% of removed and 65\% of moved objects, against 52\% and 59\% for SceneDiff.
Its false alarm rate on first-vs-second-half controls is 12\%, against 93\% for the same Qwen3.5 asked holistically.
With the same weights, the model reliably judges whether an object is still there once it knows where to look.
What a VLM lacks for change detection is the place where the object was.

\section{Method}
\label{sec:method}

\begin{figure*}[t]
\centering
\includegraphics[width=\linewidth]{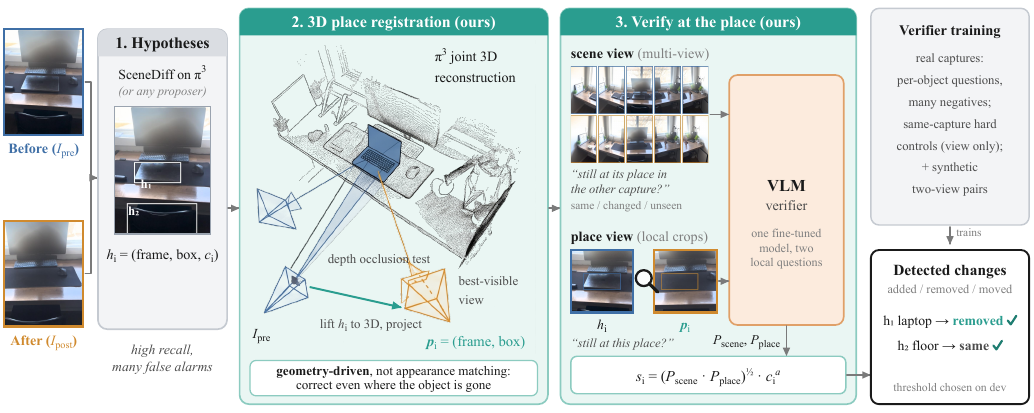}
\caption{\textbf{Overview of \ours.}
(1) A high-recall proposer, SceneDiff by default, gives change hypotheses $h_i$, each with a frame, a box and a geometric cost $c_i$.
(2) $\pi^3$ reconstructs both captures jointly. Each hypothesis is lifted to 3D and projected into the other capture, and a depth occlusion test selects the frame in which its place is best visible, which gives the registered place $p_i$. This step relies only on the static scene, so it stays correct where the object is gone.
(3) One fine-tuned VLM answers two questions. The scene view (5+5 frames) asks whether the object is still at its place in the other capture, and the place view (object crop next to place crop) asks whether it is still at this place. The fused score $s_i$ is thresholded at the operating point chosen on dev.
The verifier is trained on per-object questions of the same form, built from real captures with same-capture hard controls and from synthetic two-view pairs.}
\label{fig:pipeline}
\end{figure*}

\ours casts change detection as a set of verification questions asked at the place where each object was.
Whether an object changed is decided by the state of its place in the other capture.
Every component follows from this formulation. 3D geometry answers where to ask, a VLM answers what is there, and the supervision that trains the VLM poses the same question.
The model thus faces the same local judgement in training and at test time, and the hard part, finding correspondence across videos, is left entirely to geometry (\cref{fig:pipeline}).

\subsection{Change as a state at a place}
\label{sec:formulation}

Let the two captures of a scene be $\mathcal{A}=\{I_t^{\mathcal{A}}\}$ and $\mathcal{B}=\{I_t^{\mathcal{B}}\}$.
A change hypothesis $h=(\mathcal{X},t,b,c)$ consists of its capture $\mathcal{X}\in\{\mathcal{A},\mathcal{B}\}$, a frame index $t$, a box $b$ and a geometric cost $c$, and $\bar{\mathcal{X}}$ denotes the other capture.
For the object inside $b$, we define its place $p(h)$ in $\bar{\mathcal{X}}$ as the projection of the same 3D region into a frame of $\bar{\mathcal{X}}$.
Change detection then reduces to a three-way label $y(h)\in\{\text{same},\text{changed},\text{unseen}\}$ at $p(h)$.
The label is \emph{same} when the object is still at this place, \emph{changed} when the place is visible in $\bar{\mathcal{X}}$ without the object, and \emph{unseen} when the place is not in view in any frame of $\bar{\mathcal{X}}$.
A changed hypothesis from the earlier capture is a removed object, one from the later capture is an added object, and a moved object gives a changed label at both its old and its new place.

This formulation builds the two most common nuisances into the answer space.
Objects visible on one side only fall into unseen instead of being taken for added or removed objects.
Objects seen from a new viewpoint or under new lighting are still at $p(h)$ and receive the answer same.

The hypotheses can come from any high-recall proposer, by default the change regions of SceneDiff~\cite{wu2026scenediff} with a box and a cost per frame, or Grounding DINO~\cite{liu2024grounding} boxes with a constant $c$.

\subsection{Finding the place with geometry}
\label{sec:registration}

We reconstruct all frames of both captures jointly with $\pi^3$~\cite{wang2026pi}.
This gives per-pixel 3D points $X_t(u)$, confidences and camera-to-world poses $T_t$ for each frame, and we fit the intrinsics $K_t$ of each frame to its camera-space points by least squares.
For a hypothesis $h$, let $S(h)$ be the 3D points of the pixels inside $b$ with confidence above 0.1.
For a frame $j$ of $\bar{\mathcal{X}}$, let $\pi_j(x)$ be the projection of a point $x$ into frame $j$, $z_j(x)$ its depth in the camera of frame $j$, and $D_j$ the reconstructed depth map of frame $j$.
A point $x$ is visible in frame $j$ when
\begin{equation}
\mathrm{vis}_j(x)=\mathbb{1}\big[\,\pi_j(x)\in\Omega\;\wedge\;z_j(x)\le(1+\epsilon)\,D_j(\pi_j(x))\,\big],
\end{equation}
where $\Omega$ is the image domain and $\epsilon=0.05$.
The visible share of frame $j$ is $v_j=\frac{1}{|S(h)|}\sum_{x\in S(h)}\mathrm{vis}_j(x)$.
The place takes the frame $j^*=\argmax_j v_j$ with the largest visible share and the quantile box of the visible projected points,
\begin{equation}
p(h)=\Big(j^*,\;Q_{5,95}\big\{\pi_{j^*}(x)\;\big|\;x\in S(h),\ \mathrm{vis}_{j^*}(x)=1\big\}\Big),
\end{equation}
where $Q_{5,95}$ is the box spanned by the 5\% and 95\% quantiles of the horizontal and vertical coordinates.

The visibility test keeps the place accurate in cluttered scenes.
Points hidden behind a foreground object are not projected onto that object, so the place box lands on the tabletop or floor where the object stood, and the visible share measures how well this region is observed in the other capture.
The computation depends only on the static geometry of the scene, so the place of a removed object is recovered just as accurately, and one reconstruction serves all hypotheses of a scene.

\subsection{Asking at the place}
\label{sec:verifier}

A multi-image VLM $q_\theta$ verifies each hypothesis by answering the same question at two scales.

The scene view provides the full context.
The model sees 5 frames from each capture, including the frame of $h$, and the prompt gives the frame and box of $h$ and asks whether this object is still at its place in the other capture, with the answers same, changed or unseen.
This view lets the model judge whether the place is in view at all and separate true changes from local occlusion.

The place view provides direct local evidence.
We crop the object in the frame of $h$ and the place in the frame of $p(h)$ with $1.6\times$ context, draw the boxes on both crops, show the crops side by side, and ask whether this object is still at this place, with the answers same or changed.
When the object has been removed, the second crop shows an empty tabletop or floor, so absence becomes something the model can see.

The same weights answer both questions. With $V_{\text{scene}}(h)$ and $V_{\text{place}}(h)$ the inputs of the two views, we read the probability of changed on the first answer token,
\begin{equation}
P_{\text{v}}(h)=q_\theta\big(\text{changed}\mid V_{\text{v}}(h)\big),\qquad \text{v}\in\{\text{scene},\text{place}\},
\end{equation}
and score the hypothesis by
\begin{equation}
s(h)=\sqrt{P_{\text{scene}}(h)\,P_{\text{place}}(h)}\;\cdot\;c^{\,a}.
\end{equation}
The geometric mean requires evidence of change from both views.
The exponent $a\in\{0,0.5,1,2\}$ and the output threshold are chosen on the dev split only.
Hypotheses above the threshold form the output and keep the masks and cross-frame tracks of the proposer.

\subsection{Supervision for place questions}
\label{sec:supervision}

Holistic fine-tuning learns from a scene-level list of changes, which says nothing about where to compare.
\ours rewrites the same annotations into place questions of exactly the test-time form, one object and one place per example.
On the training videos, Grounding DINO proposes a large pool of object boxes, and each box receives a place by the same geometric procedure.
Boxes with IoU of at least 0.4 with a ground-truth changed object are labelled changed, and ambiguous boxes with IoU between 0.1 and 0.4 are discarded.
The remaining boxes are labelled same or unseen depending on whether their region is observed in the other capture.
Each box gives one scene-view and one place-view question.
Changed questions are duplicated, and negatives are drawn in equal parts from real training items and control pairs.

Control pairs give the most informative negatives. In a first-vs-second-half control, many objects appear on one side only, exactly where holistic VLMs fail, and all of them are labelled same or unseen.
So that the same model also handles two-image input, we add two-frame versions of the SceneDiff training questions and questions built from the synthetic two-view dataset KC-3D of CYWS-3D~\cite{sachdeva2023change}, where ground-truth masks give the changes and $\pi^3$ registers the places in the same way.

The training set holds about 112k questions, and we fine-tune all parameters of Qwen3.5-9B for one epoch with the loss on the answer token.
Training and testing share the same registration and views, so at test time the verifier meets its training question on new scenes and new proposal sources.

\section{Experiments}
\label{sec:experiments}

\begin{table*}[t]
\centering
\caption{\textbf{Results under the official SceneDiff protocol} (test split, mask IoU 0.5).
obj/sc AP, the main metric of the benchmark, scores each changed object once per scene over all frames of both videos.
Rows above the line are from~\cite{wu2026scenediff}, and $^\ast$ marks fine-tuning on the SceneDiff dev split.
\ours{} verifies the object hypotheses of SceneDiff and keeps their masks, and px/im IoU drops hypotheses below the operating point chosen on SD-V dev.
Qwen3.5-9B: mean$\pm$std over three training runs. SD-K uses the SD-V operating point without any kitchen data.
Bold: best. Underlined: second best.}
\label{tab:official}
\setlength{\tabcolsep}{5pt}
\begin{adjustbox}{max width=\linewidth}
\begin{tabular}{@{}llcccccc@{}}
\toprule
& & \multicolumn{3}{c}{SD-V (in-domain)} & \multicolumn{3}{c}{SD-K (cross-dataset)} \\
\cmidrule(lr){3-5}\cmidrule(l){6-8}
Method & Output & px/im IoU & obj/im AP & obj/sc AP & px/im IoU & obj/im AP & obj/sc AP \\
\midrule
MV3DCD~\cite{galappaththige2025multi} & pixel & 22.1 & -- & -- & 9.1 & -- & -- \\
3DGS-CD~\cite{lu20253dgs} & track & 14.7 & 1.0 & 0.5 & 4.8 & 0.1 & 0.1 \\
CYWS-3D~\cite{sachdeva2023change} & box & -- & 13.9 & -- & -- & 6.6 & -- \\
CYWS-2D~\cite{sachdeva2023changewacv} & box & -- & 19.4 & -- & -- & 9.0 & -- \\
CYWS-2D$^\ast$~\cite{sachdeva2023changewacv} & box & -- & 24.5 & -- & -- & 16.8 & -- \\
VLM: Qwen2.5-VL + SAM3~\cite{wu2026scenediff} & track & 22.9 & 7.1 & 5.3 & 12.4 & 3.6 & 2.1 \\
SceneDiff~\cite{wu2026scenediff} & track & 39.1 & 43.8 & 22.8 & 20.8 & 20.9 & 10.6 \\
\midrule
SceneDiff, reproduced & track & 37.1 & 44.4 & 24.0 & 20.4 & 21.6 & 11.6 \\
\rowcolor{oursbg}\ours{} (Qwen3-VL-8B) & track & \second{46.4} & \second{50.5} & \second{36.7} & \second{35.1} & \best{26.6} & \best{21.8} \\
\rowcolor{oursbg}\ours{} (Qwen3.5-9B) & track & \best{48.6\std{0.8}} & \best{51.0\std{0.7}} & \best{37.8\std{0.3}} & \best{40.0\std{0.8}} & \second{26.1\std{0.1}} & \second{20.6\std{1.2}} \\
\multicolumn{2}{l}{\footnotesize\textit{vs.\ reproduced SceneDiff}} & \gain{+11.5} & \gain{+6.6} & \gain{+13.8} & \gain{+19.6} & \gain{+4.5} & \gain{+9.0} \\
\bottomrule
\end{tabular}
\end{adjustbox}
\end{table*}

\subsection{Setup}
\label{sec:setup}

\paragraph{Data.}
We evaluate on the two subsets of SceneDiff~\cite{wu2026scenediff} and on the two-view benchmark RC-3D~\cite{sachdeva2023change}.
SD-V contains 200 video pairs of everyday scenes in homes, stores, offices and outdoors.
We use only the 50 pairs of its dev split, 40 for training and 10 for choosing the operating point, and never touch the 150 test pairs during training or tuning.
SD-K contains 150 egocentric kitchen video pairs from HD-EPIC, and we evaluate directly on its 100 test pairs with no kitchen data and no new operating point.
RC-3D contains 100 real two-view image pairs with one changed object each and is used for testing only.

\paragraph{Protocols.}
The official SceneDiff code reports px/im IoU, per-frame obj/im AP and the main metric, scene-level obj/sc AP.
Here \ours verifies each object detected by SceneDiff and keeps its masks and tracks, so every change in the metrics comes from the verification.
For a fair comparison with holistic VLMs, we also use a VLM-output protocol in their format, with the 5+5-frame items and controls of \cref{sec:diagnosis}, a hit at IoU$\geq$0.3, and AP and F1 at the operating point.
On RC-3D we report AP50 as in SceneDiff, with the absent-view boxes of all methods from the same $\pi^3$ projection.

\paragraph{Baselines.}
We compare with all methods reported by SceneDiff~\cite{galappaththige2025multi,lu20253dgs,sachdeva2023changewacv,sachdeva2023change} and with our reproduction of SceneDiff (24.0 obj/sc AP on SD-V against 22.8 in the original paper).
Holistic VLM baselines are zero-shot Qwen3.5-9B, Qwen3-VL-8B~\cite{bai2025qwen3} and InternVL3.5-8B~\cite{wang2025internvl3}, and Qwen3.5-9B and Qwen3-VL-8B fine-tuned on the training videos and annotations of \ours to output the change list directly.

\paragraph{Implementation.}
Training takes about 9 hours on 4 A100 GPUs at learning rate $2\times10^{-6}$.
We train the main system with three seeds, choose the operating point on the 10 SD-V dev pairs only, and compute all confidence intervals with a scene bootstrap.

\subsection{Official protocol}
\label{sec:official}

\cref{tab:official} reports the official SceneDiff protocol.
\ours raises obj/sc AP from 24.0 to 37.8 on SD-V and from 11.6 to 20.6 on SD-K, with standard deviations of 0.3 and 1.2 across the three training runs.
Per-frame obj/im AP rises from 44.4 to 51.0 and from 21.6 to 26.1, and px/im IoU from 37.1 to 48.6 and from 20.4 to 40.0.
When the change type must also be correct, obj/sc AP rises from 19.5 to 30.1 and from 8.4 to 13.7.
\ours keeps the objects and masks of SceneDiff, so all of these gains come from judging more accurately whether each candidate object actually changed.
A Qwen3-VL-8B verifier performs on par (36.7 and 21.8).

\subsection{Comparison with VLM baselines}
\label{sec:vlm}

\begin{table}[t]
\centering
\caption{\textbf{Comparison with VLM baselines} under the VLM-output protocol (5+5 frames, a list of changes with frame and box, IoU 0.3).
F1 is taken at the operating point chosen on SD-V dev, and SD-K uses the same point.
Spec: share of first-vs-second-half controls with no reported change.
$^\dagger$Boxes regenerated with the model's native grounding prompt. $\times$3: mean of three training runs.}
\label{tab:vlm}
\setlength{\tabcolsep}{3.2pt}
\footnotesize
\begin{adjustbox}{max width=\linewidth}
\begin{tabular}{@{}lccccc@{}}
\toprule
& \multicolumn{2}{c}{SD-V} & \multicolumn{2}{c}{SD-K} & \\
\cmidrule(lr){2-3}\cmidrule(lr){4-5}
Method & AP & F1 & AP & F1 & Spec \\
\midrule
\multicolumn{6}{@{}l}{\textit{Zero-shot multi-image VLM}} \\
\quad Qwen3.5-9B & 1.2 & 9.2 & 0.2 & 3.9 & -- \\
\quad Qwen3-VL-8B & 2.8 & 14.3 & 1.3 & 9.6 & 0.4 \\
\quad InternVL3.5-8B$^\dagger$ & 2.9 & 14.1 & 1.1 & 8.7 & -- \\
\midrule
\multicolumn{6}{@{}l}{\textit{Fine-tuned on the same data}} \\
\quad Qwen3-VL-8B & 15.8 & 33.1 & 7.6 & 24.3 & -- \\
\quad Qwen3.5-9B ($\times$3) & 19.4 & 39.2 & 9.1 & 26.8 & 30.2 \\
\midrule
\multicolumn{6}{@{}l}{\textit{Geometry, training-free}} \\
\quad SceneDiff~\cite{wu2026scenediff} & 25.6 & 42.9 & 22.5 & 33.5 & \second{91.7} \\
\midrule
\multicolumn{6}{@{}l}{\textit{Registered two-view verification}} \\
\quad \ours{}, zero-shot Qwen3.5-9B & 35.6 & 45.2 & 38.6 & 46.9 & 87.9 \\
\rowcolor{oursbg}\quad \ours{} (Qwen3-VL-8B) & \second{55.5} & \second{59.6} & \best{47.0} & \second{49.5} & -- \\
\rowcolor{oursbg}\quad \ours{} (Qwen3.5-9B) ($\times$3) & \best{57.6} & \best{61.1} & \second{46.9} & \best{52.7} & \best{92.7} \\
\bottomrule
\end{tabular}
\end{adjustbox}
\end{table}

\cref{tab:vlm} compares all methods under the VLM-output protocol.
Zero-shot VLMs stay below 3 AP.
\ours reaches 57.6 AP and 61.1 F1 on SD-V, 21.9 F1 points above same-data holistic fine-tuning (95\% interval $[15, 28]$) and 18.2 above SceneDiff ($[11, 26]$).
On SD-K its F1 of 52.7 exceeds the two baselines by 25.9 and 19.2 points, with both lower interval bounds above 14.
On first-vs-second-half controls \ours stays silent on 92.7\% of the pairs, against 30.2\% for the model fine-tuned on the same data.

The gains have two main sources.
Compared with SceneDiff, \ours raises precision from 34\% to 62\% and recall slightly from 58\% to 60\%, because the verifier rejects the many false alarms that lighting, clutter and viewpoint changes trigger in the geometric detector.
The largest improvement is in cluttered public indoor scenes such as stores, offices and gyms, where F1 rises from 29 to 52.
Compared with holistic fine-tuning, recall on removed and moved objects rises from 36\% and 34\% to 62\% and 58\%, exactly the two types that require looking at the place where the object was.

Stronger holistic baselines stay far below \ours.
Adding first-vs-second-half controls to holistic fine-tuning raises the specificity on them to 83.7\% but lowers F1 to 37.1, and adding all per-object questions of \ours as an auxiliary task gives 40.5 F1.
The same annotations yield the gains only when the question is asked at the registered place.

\begin{figure*}[t]
\centering
\includegraphics[width=0.92\linewidth]{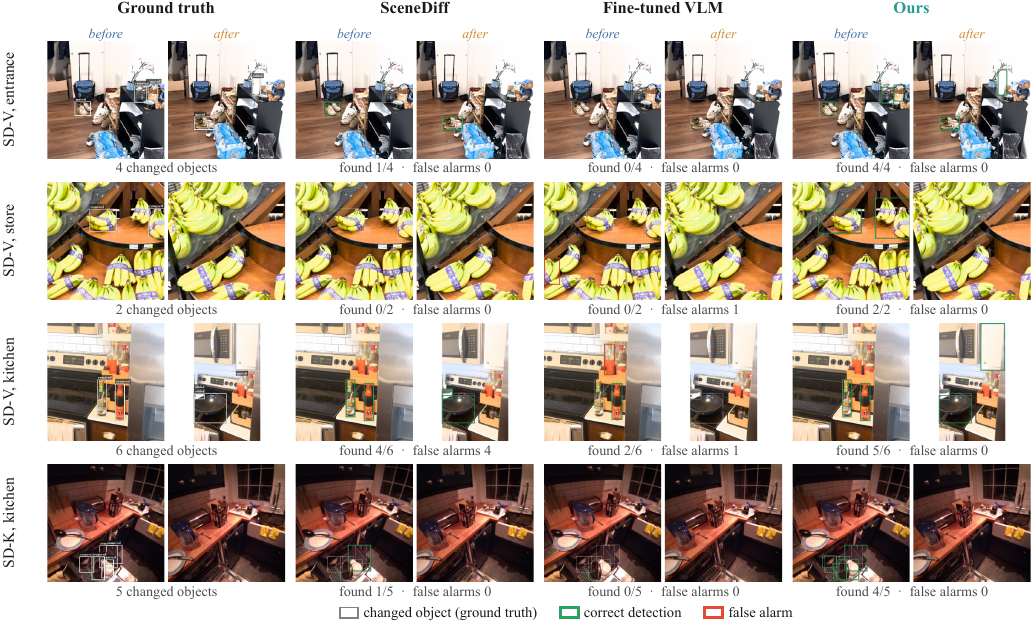}
\caption{\textbf{Qualitative comparison} on three SD-V scenes and one SD-K scene. Each row shows two of the 5+5 frames of an item, and the counts are over all frames.
Gray boxes: ground-truth changed objects. Green: correct detections. Red: false alarms.
Over the four items, SceneDiff finds 6 of 17 changes with 4 false alarms, the fine-tuned VLM finds 2 with 2 false alarms, and \ours finds 15 with none.}
\label{fig:qual}
\end{figure*}

\subsection{What the place view needs}
\label{sec:ablation}

\begin{table}[t]
\centering
\caption{\textbf{What the place view needs} (SD-V test split).
Same verifier, only the place view changes, operating point chosen on dev per row. Coverage: share of hypotheses that receive a place. Brackets: difference to the $\pi^3$ place view, every 95\% scene-bootstrap interval excluding zero.}
\label{tab:ablation}
\footnotesize
\setlength{\tabcolsep}{4pt}
\begin{adjustbox}{max width=\linewidth}
\begin{tabular}{@{}lccc@{}}
\toprule
Place view & Coverage & AP & F1 \\
\midrule
\rowcolor{oursbg}$\pi^3$-registered place (ours) & 99\% & \best{59.2} & \best{62.4} \\
SIFT-matched place & 59\% & 56.4\,\dlt{\text{-}2.8} & 57.9\,\dlt{\text{-}4.5} \\
$\pi^3$ frame, box not mapped & 99\% & 52.2\,\dlt{\text{-}7.0} & 55.0\,\dlt{\text{-}7.4} \\
no place view (scene view only) & -- & 53.8\,\dlt{\text{-}5.4} & 54.3\,\dlt{\text{-}8.0} \\
place view only (no scene view) & 99\% & 50.9\,\dlt{\text{-}8.3} & 50.4\,\dlt{\text{-}12.0} \\
\bottomrule
\end{tabular}
\end{adjustbox}
\end{table}

\cref{tab:ablation} changes only the place view.
With SIFT matching instead of $\pi^3$ registration, only 59\% of the hypotheses obtain a place, and AP drops by 2.8 and F1 by 4.5.
The hypotheses that SIFT cannot place are mostly changed objects, because the object to be matched is no longer there.
Taking the frame chosen by $\pi^3$ but keeping the box at its original image coordinates drops AP by 7.0, so both choosing the right frame and mapping the box onto the place matter.
Removing the place view and keeping only the scene view drops AP by 5.4, and keeping only the place view drops it by 8.3.
The two views provide complementary evidence.

\subsection{Generality}
\label{sec:general}

\begin{table}[t]
\centering
\caption{\textbf{Generality across proposers, verifiers and benchmarks.}
The verifier works with change-agnostic proposals and gains from change-aware ones, registered verification helps even an untrained VLM, and a second VLM family behaves alike. RC-3D: AP50 with the $\pi^3$ projection for all rows. Shaded: main system (three runs).}
\label{tab:general}
\begin{adjustbox}{max width=\linewidth}
\begin{tabular}{@{}lcccc@{}}
\toprule
& \multicolumn{2}{c}{SD-V} & SD-K & RC-3D \\
\cmidrule(lr){2-3}\cmidrule(lr){4-4}\cmidrule(l){5-5}
Setting & AP & F1 & F1 & AP50 \\
\midrule
\multicolumn{5}{@{}l}{\textit{Proposals (same verifier)}} \\
\quad Grounding DINO, change-agnostic & 43.0 & 48.6 & -- & -- \\
\quad SceneDiff change regions & 57.4 & 60.8 & -- & -- \\
\midrule
\multicolumn{5}{@{}l}{\textit{Verifier}} \\
\quad untrained Qwen3.5-9B & 35.6 & 45.2 & 46.9 & 61.0 \\
\quad trained, multi-view data only & 59.9 & 61.3 & 50.6 & 63.7 \\
\rowcolor{oursbg}\quad trained, + synthetic two-view pairs & 57.6 & 61.1 & 52.7 & 64.3 \\
\quad trained, Qwen3-VL-8B backbone & 55.5 & 59.6 & 49.5 & 63.9 \\
\midrule
\multicolumn{5}{@{}l}{\textit{Reference}} \\
\quad SceneDiff~\cite{wu2026scenediff}, reproduced & 25.6 & 42.9 & 33.5 & 61.0 \\
\bottomrule
\end{tabular}
\end{adjustbox}
\end{table}

With change-agnostic Grounding DINO proposals (\cref{tab:general}), the same verifier reaches 43.0 AP on SD-V, already well above SceneDiff (25.6) and holistic fine-tuning (19.4).
With SceneDiff change regions, AP rises further to 57.4, as high-recall geometric proposals and verification at the registered place complement each other.
Training lifts the verifier from 45.2 F1 (untrained Qwen3.5) to 61.1, and a Qwen3-VL-8B verifier reaches 59.6 F1, 26.5 points above holistic fine-tuning of that backbone.

Synthetic two-view pairs bring the same model to 64.3 AP50 on RC-3D, 3.3 points above SceneDiff under the same projection (95\% interval $[0.2, 6.2]$), with unchanged multi-view F1.
A perfect verifier on the SceneDiff candidates reaches 70.7 AP50 on RC-3D and 74.7 AP on SD-V, so \ours captures about a third of this headroom on RC-3D and about 65\% on SD-V.
Registration takes under 1 second and verification about 2.7 seconds per item, only about 11\% on top of the 25 seconds of SceneDiff.

\subsection{Qualitative results}
\label{sec:qual}

In the four scenes of \cref{fig:qual}, \ours finds all changes in the home entrance and the store shelf, where the fine-tuned VLM finds none, and still finds 4 of 5 small, densely packed objects in the egocentric SD-K kitchen.

\section{Can a VLM Find the Place Itself?}
\label{sec:learned}

\ours leaves finding the place to geometry.
A natural question is whether a VLM can learn this step as well, so that the whole system relies on one model.
We trained a series of localisation models, supervised by the $\pi^3$ places of the training videos, that take a hypothesis in the scene view and output a frame and a box in the other capture.
On top of the base version we added zoomed refinement, more weight on changed objects with extra synthetic data, query crops with wider context and a second refinement pass.
We also optimised localisation with reinforcement learning, rewarding the IoU with the $\pi^3$ place and the probability that the verifier gives to the correct answer at the predicted place.
To keep this reward informative on unseen real scenes, we used cross-fitting, with supervised fine-tuning on one half of the training videos and reinforcement learning on the other.

\begin{figure}[t]
\centering
\includegraphics[width=0.66\linewidth]{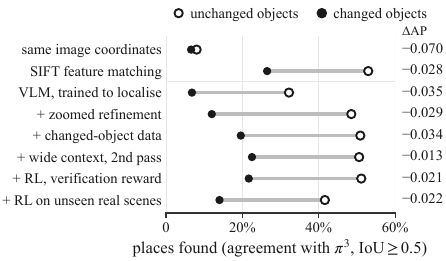}
\caption{\textbf{The registration paradox.} Places found (same frame as the $\pi^3$ place, IoU$\geq$0.5) for unchanged (open) and changed (filled) objects by the box left in place, SIFT matching and successive versions of learned VLM localisation. Appearance-based registration finds unchanged places about twice as often as changed ones, where the answer is decided. Right: system AP change against $\pi^3$ places.}
\label{fig:registration}
\end{figure}

We count a place as found when the prediction lies in the same frame as the $\pi^3$ place with IoU of at least 0.5, separately for unchanged and changed objects (\cref{fig:registration}).
In every version, the places of unchanged objects are found 32\% to 52\% of the time, those of changed objects only 7\% to 23\% of the time.
SIFT matching shows nearly the same split, with 53\% on unchanged and 26\% on changed objects.
Reinforcement learning raises the hit rate on changed objects from 27\% to 63\% on its training scenes, yet it stays at 14\% on the test scenes.
With these learned places, the same verifier yields 1.3 to 3.5 points lower system AP than with $\pi^3$ places.

The pattern reveals a structural limit of appearance matching.
Feature matching and learned VLM localisation both rely on the appearance of the object to find its correspondence, and where an object was removed or moved this appearance no longer exists in the other capture.
The model must then infer the place from the surroundings, which is far harder under new viewpoints and does not generalise from a few training scenes to new ones.
3D reconstruction solves for cameras and geometry from the static structure of the whole scene, whether or not any single object is still there.
The formulation of \ours therefore takes the place from geometry and asks the VLM only for the judgement at that place.

\section{Conclusion}
\label{sec:conclusion}

Multi-image VLMs fail to detect scene changes mainly because they cannot find where an object was, and appearance matching cannot recover this place once the object is gone.
\ours reformulates change detection as object verification at a registered place, where 3D geometry supplies the place and a VLM, trained with supervision of the same form, judges from a scene view and a place view whether the object is still there.
\ours improves obj/sc AP by 57\% and 78\% on the two SceneDiff subsets, improves two-view RC-3D, and transfers across proposers and backbones.

\paragraph{Limitations.}
Recall is bounded by the proposer and place accuracy by the 3D reconstruction, and \ours can adopt stronger models for both (see \cref{sec:limits}).

\bibliographystyle{unsrtnat}
\bibliography{ref}

\clearpage
\label{sec:suppl-start}

\appendix
\renewcommand{\thefigure}{S\arabic{figure}}
\renewcommand{\thetable}{S\arabic{table}}
\renewcommand{\theequation}{S\arabic{equation}}
\setcounter{figure}{0}
\setcounter{table}{0}
\setcounter{equation}{0}

\section*{Appendix}
\noindent\textbf{Contents}\par\smallskip
{\small
\noindent\begin{tabular}{@{}l@{\hspace{0.6em}}l@{}}
\ref{sec:impl} & Implementation details \\
\ref{sec:protocols} & Evaluation protocols \\
\ref{sec:full} & Full results \\
\ref{sec:moreabl} & Additional ablations \\
\ref{sec:analysis} & Result analysis \\
\ref{sec:learnedfull} & Full record of learned registration \\
\ref{sec:cost} & Inference cost \\
\ref{sec:morequal} & Qualitative results \\
\ref{sec:dense} & Relation to dense change detection benchmarks \\
\ref{sec:limits} & Limitations and outlook \\
\end{tabular}}

\section{Implementation details}
\label{sec:impl}

\paragraph{Prompts.}
The scene view gives the frames of the two captures in order, each preceded by a text header such as ``Video 1, frame 3:''.
The frames are followed by the prompt below, where \{v\}, \{f\} and \{b\} are the video, frame and box of the hypothesis and \{w\} is the other video.

\begin{quote}\small\ttfamily
Look at the object in video \{v\}, frame \{f\}, inside the box \{b\} (0-1000 coordinates). Compare it with the frames of video \{w\}. Answer "same" if this object is still at its place in video \{w\}, "changed" if it is not there any more (it was removed or moved away; for an object of video 2: it was not there before), or "unseen" if its place is not in view in any frame of video \{w\}. Answer with one word.
\end{quote}

\noindent The place view gives the object crop and then the place crop, with the region marked by a red box on both images, followed by the prompt below.

\begin{quote}\small\ttfamily
Image 1 shows an object (inside the red box) in one capture of a scene. Image 2 shows the same place, marked by the red box, in the other capture. Answer "same" if this object is still at this place in image 2, or "changed" if it is not there (it was removed or moved away; if image 1 is from the later capture: it was not there before). Answer with one word.
\end{quote}

\noindent The holistic VLM baseline receives the same 5+5 frames and the prompt below.

\begin{quote}\small\ttfamily
Between time 1 and time 2 some objects in the scene may have been added, removed or moved. List every such object. Ignore differences that come only from the camera position, framing or lighting. Answer with a JSON list. Each entry is \{"change": "added" | "removed" | "moved", "label": "<object>", "video": 1 or 2, "frame": <frame number>, "bbox\_2d": [x1, y1, x2, y2]\}: one frame in which the object is visible (a frame of time 1 for a removed object, of time 2 for an added object, either for a moved object) and its box in that frame in 0-1000 coordinates. Answer [] if nothing changed.
\end{quote}

\paragraph{Training data.}
The verifier training set contains 112,466 examples from five sources.
The first is 16,336 scene-view questions on the SD-V training videos, built from Grounding DINO object boxes with confidence of at least 0.25.
A box is labelled changed when its IoU with a ground-truth changed object is at least 0.4 and discarded when this IoU lies between 0.1 and 0.4.
Otherwise it is labelled same when it has at least 5 SIFT correspondences in the other capture and unseen when it has fewer.
This rule only generates training labels.
Changed questions are repeated twice, and same and unseen questions are drawn in equal parts from real training items and from control pairs of both kinds.
The second source is the 13,803 place-view crop pairs of the same questions, with places registered by $\pi^3$.
The third is 13,803 two-frame SD-V questions for two-view input, which keep only the frame of the candidate and its $\pi^3$ place frame.
Here the place is always in view, so unseen merges into same.
The fourth is 36,252 questions from 2,974 image pairs of the synthetic two-view dataset KC-3D.
Connected components of the change masks enter directly as changed questions, Grounding DINO boxes with IoU below 0.1 to every change region are labelled same, and same questions are capped at three times the changed ones.
The fifth is the 32,272 KC-3D questions that obtain a $\pi^3$ place, which give the corresponding place-view questions.

\paragraph{Training and inference.}
The verifier fine-tunes all parameters of Qwen3.5-9B with FSDP on 4 A100 80GB GPUs.
We use learning rate $2\times10^{-6}$ with cosine decay, 3\% warm-up and an effective batch size of 16, and train one epoch of about 7,000 steps in 9 hours with cross-entropy on the answer token.
The pixel budget is $256\times1024$ per frame in the scene view, and each place-view crop has a longest side of 384 pixels.
At inference we read the probabilities of same, changed and unseen on the first answer token without generation, and the questions of one item share a prefix cache.
The Qwen3-VL-8B version uses exactly the same data and settings.

\paragraph{Place registration.}
$\pi^3$ reconstructs all frames of a test item jointly once, at an input resolution of $518\times294$.
The intrinsics are fitted by least squares to the camera-space points with confidence above 0.1, separately for $u=f_x X/Z+c_x$ and $v=f_y Y/Z+c_y$.
The visibility test uses a depth tolerance of 5\%, the place box spans the 5\% and 95\% quantiles of the visible projected points, and the place crop enlarges the box by $1.6\times$ with a shortest side of at least 15\% of the image side.
Of the 15,776 hypotheses of the SD-V test split, 99\% obtain a place.

\paragraph{Operating point and statistics.}
The fusion exponent $a\in\{0,0.5,1,2\}$ and a threshold among 60 quantiles of the dev scores are chosen jointly to maximise F1 on the SD-V dev split.
The three training runs select $a=1$ with threshold 0.0031, $a=2$ with 0.0018 and $a=2$ with 0.0012, and these points are used unchanged on SD-K and RC-3D.
All confidence intervals come from 2,000 bootstrap resamples over scenes, with 150 scenes on SD-V, 100 scenes on SD-K and 100 image pairs on RC-3D.
When the three-run system is compared with a baseline, each run uses its own operating point.

\section{Evaluation protocols}
\label{sec:protocols}

\paragraph{VLM-output protocol.}
Each test video pair gives 6 test items, namely one real item with uniformly sampled frames, two real items with randomly sampled frames, and the three corresponding controls, for 900 items on SD-V and 600 on SD-K.
In real items every changed object has a box in every frame where it is visible.
Objects whose largest visible area is below 0.3\% of the image are ignore regions, and predictions on them count neither as correct nor as false.
About 10\% of the raw SceneDiff videos were recorded at 60 or 120 frames per second, and we convert frame indices by the ratio of frame counts between the annotated and the raw video.
A prediction hits an object when the object is visible in the predicted frame and the box IoU is at least 0.3, and each object counts once.
A further 289 first-vs-second-half controls serve only to count false alarms.

\paragraph{Verification under the official protocol.}
Following the official code, we extract the test frames at 1 frame per second with ffmpeg, and our reproduction of SceneDiff matches the original paper within 0.5 to 2 points on every metric of SD-V and SD-K.
For each object detected by SceneDiff, we build a hypothesis in every frame where it has a mask, take the scene view from the 5+5 frames around that frame, and replace the cost of the frame by the fused score.
Masks and cross-frame tracks stay unchanged, and the official code averages the frame scores into the object score.
px/im IoU evaluates the union of all output masks, so for this metric we additionally drop the objects whose mean score is below the dev threshold.

\paragraph{RC-3D.}
RC-3D requires a box both in the view where the object is present and in the view where it is absent.
SceneDiff outputs masks only in the view where the object is present, so we project the box into the absent view with $\pi^3$.
This projection is identical for SceneDiff and \ours.
The SceneDiff paper obtains the absent-view box by lifting the mask pixels to 3D and projecting them, and reports 65.3 AP50 over both views, while our reproduction under the shared box projection reaches 61.0.

\section{Full results}
\label{sec:full}

\begin{table*}[t]
\centering
\caption{\textbf{Every training run of the main system} (Qwen3.5-9B, three seeds). Each run uses the fusion exponent and threshold chosen on its own SD-V dev split, and SD-K and RC-3D use that operating point unchanged.
HC spec: specificity on first-vs-second-half controls. obj/sc$^c$: the change type must also be correct.}
\label{tab:runs}
\footnotesize
\setlength{\tabcolsep}{4pt}
\begin{adjustbox}{max width=\linewidth}
\begin{tabular}{@{}lcccccccccc@{}}
\toprule
& \multicolumn{5}{c}{our protocol} & RC-3D & \multicolumn{4}{c}{official protocol} \\
\cmidrule(lr){2-6}\cmidrule(lr){7-7}\cmidrule(l){8-11}
& SD-V AP & SD-V F1 & SD-K AP & SD-K F1 & HC spec & AP50 & obj/sc SD-V & obj/sc SD-K & obj/sc$^c$ SD-V & obj/sc$^c$ SD-K \\
\midrule
run 0 (seed 0) & 59.2 & 62.4 & 45.3 & 51.5 & 92.4 & 65.8 & 38.1 & 19.4 & 30.1 & 12.8 \\
run 1 (seed 1) & 56.0 & 59.4 & 47.7 & 53.0 & 91.7 & 64.0 & 37.7 & 21.8 & 29.8 & 14.9 \\
run 2 (seed 2) & 57.5 & 61.5 & 47.7 & 53.4 & 94.1 & 63.1 & 37.5 & 20.5 & 30.4 & 13.6 \\
\midrule
mean$\pm$std & 57.6\std{1.6} & 61.1\std{1.6} & 46.9\std{1.4} & 52.7\std{1.0} & 92.7\std{1.2} & 64.3\std{1.4} & 37.8\std{0.3} & 20.6\std{1.2} & 30.1\std{0.3} & 13.7\std{1.0} \\
\bottomrule
\end{tabular}
\end{adjustbox}
\end{table*}

\begin{figure}[t]
\centering
\includegraphics[width=0.7\linewidth]{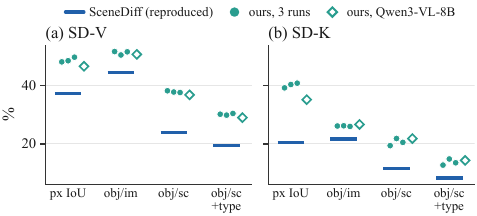}
\caption{\textbf{Official protocol, every run.} Horizontal bars: reproduced SceneDiff. Dots: the three Qwen3.5-9B runs of \ours. Diamonds: the Qwen3-VL-8B run. obj/sc+type also requires the correct change type.}
\label{fig:official}
\end{figure}

\paragraph{Per-run results.}
\cref{tab:runs} lists all results of the three training runs under both protocols and on RC-3D.
The variation across runs is far smaller than the margin over the baselines, with standard deviations of official obj/sc AP of 0.3 on SD-V and 1.2 on SD-K.
\cref{fig:official} plots every official metric for each run.

\begin{figure}[t]
\centering
\includegraphics[width=0.7\linewidth]{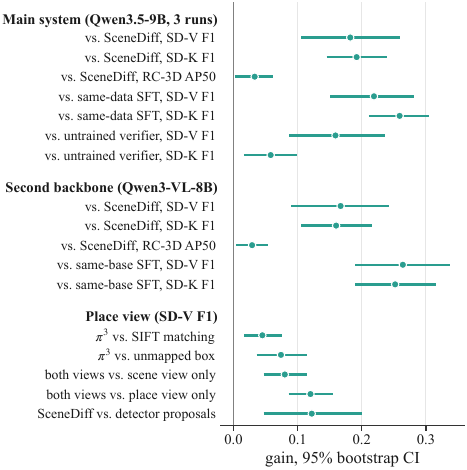}
\caption{\textbf{Every paired comparison} of the main text as the gain with its 95\% scene-bootstrap interval on a 0 to 1 scale (three runs, operating points chosen on dev). Groups: the main system, the second backbone, and the place-view ablations on SD-V F1.}
\label{fig:forest}
\end{figure}

\begin{figure}[t]
\centering
\includegraphics[width=0.59\linewidth]{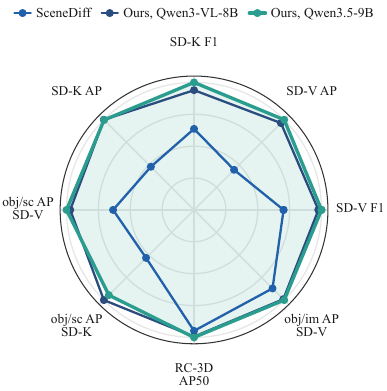}
\caption{\textbf{All metrics at a glance}, each axis scaled by its best value. Both backbones of \ours enclose SceneDiff on every axis.}
\label{fig:radar}
\end{figure}

\paragraph{Confidence intervals for all comparisons.}
\cref{fig:forest} collects the differences and 95\% intervals of every paired comparison in the main text.
They cover the comparisons with SceneDiff, with same-data holistic fine-tuning and with the untrained verifier, the comparisons on the second backbone, and the place-view ablations.
Every interval lies above zero.
\cref{fig:radar} compares the two backbones of \ours with SceneDiff on all metrics in one radar plot.

\begin{figure*}[t]
\centering
\includegraphics[width=0.75\linewidth]{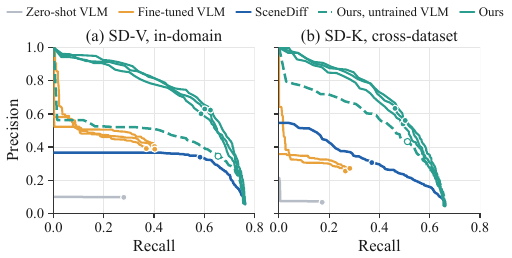}
\caption{\textbf{Precision-recall curves} under the VLM-output protocol (IoU 0.3) on SD-V (a) and SD-K (b). The three runs of \ours lie above all baselines over the whole curve. Dots: operating points chosen on SD-V dev.}
\label{fig:pr}
\end{figure*}

\paragraph{Precision-recall curves.}
\cref{fig:pr} shows the precision-recall curves on SD-V and SD-K.
All three runs of \ours lie above SceneDiff and the holistic VLMs over the whole curve on both subsets.

\paragraph{Full RC-3D results.}
\cref{tab:rc3d} reports AP50 on RC-3D in the present view, in the absent view and over both, together with the upper bound of 70.7 reached by a perfect verifier that keeps only the candidates on changed objects.

\begin{table}[t]
\centering
\caption{\textbf{Two-view change detection on RC-3D}~\cite{sachdeva2023change} (AP50). A detection needs a box in the view where the object is present and in the view where it is absent.
$^\ast$Sensor depth. $^\ddagger$Run here with the same proposals and the same $\pi^3$ projection into the absent view. The perfect verifier keeps exactly the proposals on the changed object.}
\label{tab:rc3d}
\footnotesize
\setlength{\tabcolsep}{4pt}
\begin{adjustbox}{max width=\linewidth}
\begin{tabular}{@{}lccc@{}}
\toprule
Method & Both & Present & Absent \\
\midrule
CYWS-2D~\cite{sachdeva2023changewacv} & 14.0 & -- & -- \\
CYWS-3D~\cite{sachdeva2023change} & 41.0 & -- & -- \\
CYWS-3D$^\ast$~\cite{sachdeva2023change} & 50.0 & -- & -- \\
SceneDiff~\cite{wu2026scenediff} (paper) & 65.3 & 72.3 & 59.1 \\
\midrule
SceneDiff, reproduced$^\ddagger$ & 61.0 & 71.6 & 52.5 \\
\ours{}, zero-shot verifier$^\ddagger$ & 61.0 & 71.5 & 51.8 \\
\rowcolor{oursbg}\ours{} (Qwen3-VL-8B)$^\ddagger$ & 63.9 & 75.3 & 54.3 \\
\rowcolor{oursbg}\ours{} (Qwen3.5-9B)$^\ddagger$ & 64.3\std{1.4} & 75.0\std{1.5} & 55.2\std{1.1} \\
\midrule
\textcolor{black!55}{perfect verifier$^\ddagger$} & \textcolor{black!55}{70.7} & \textcolor{black!55}{82.6} & \textcolor{black!55}{59.7} \\
\bottomrule
\end{tabular}
\end{adjustbox}
\end{table}

\section{Additional ablations}
\label{sec:moreabl}

\begin{table}[t]
\centering
\caption{\textbf{Score fusion} (SD-V test split, run 0). The verifier carries the system, and multiplying by the SceneDiff cost adds about one F1 point and a little hard-control specificity. Shaded: the exponent chosen on dev.}
\label{tab:fusion}
\footnotesize
\begin{adjustbox}{max width=\linewidth}
\begin{tabular}{@{}lccc@{}}
\toprule
Score & AP & F1 & HC spec \\
\midrule
SceneDiff cost only (= SceneDiff) & 25.6 & 42.9 & -- \\
verifier only ($a=0$) & 59.0 & 61.4 & 91.4 \\
verifier $\times$ cost$^{0.5}$ & 59.2 & 61.8 & 91.7 \\
\rowcolor{oursbg}verifier $\times$ cost$^{1}$ (dev choice) & 59.2 & 62.4 & 92.4 \\
verifier $\times$ cost$^{2}$ & 58.9 & 62.2 & 93.1 \\
$\min$(verifier, cost) & 59.0 & 61.4 & -- \\
\bottomrule
\end{tabular}
\end{adjustbox}
\end{table}

\paragraph{Score fusion.}
\cref{tab:fusion} compares fusion rules.
The verifier probability alone already gives 61.4 F1, and multiplying by the geometric cost raises F1 to 62.4 and slightly improves the specificity on first-vs-second-half controls.
The verifier judgement drives the performance of the system, and the geometric cost mainly serves to propose hypotheses.

\begin{table}[t]
\centering
\caption{\textbf{Stronger holistic baselines} (Qwen3.5-9B, SD-V test split). Adding hard controls to training, or all per-object questions of \ours{} as auxiliary tasks, raises hard-control specificity but not detection AP or F1.}
\label{tab:sft}
\footnotesize
\setlength{\tabcolsep}{3pt}
\begin{adjustbox}{max width=\linewidth}
\begin{tabular}{@{}lccc@{}}
\toprule
Holistic fine-tuning & AP & F1 & HC spec \\
\midrule
plain holistic SFT ($\times$3) & 19.4\std{1.7} & 39.2\std{1.2} & 30.2\std{5.2} \\
+ same-capture hard controls in training ($\times$3) & 17.7\std{1.0} & 37.1\std{0.6} & 83.7\std{4.8} \\
+ all per-object verifier queries as auxiliary tasks ($\times$2) & 20.9\std{1.2} & 40.5\std{0.4} & 80.1\std{4.6} \\
\midrule
\rowcolor{oursbg}\ours{} ($\times$3) & 57.6\std{1.6} & 61.1\std{1.6} & 92.7\std{1.2} \\
\bottomrule
\end{tabular}
\end{adjustbox}
\end{table}

\paragraph{Stronger holistic fine-tuning.}
\cref{tab:sft} compares three holistic fine-tuning variants.
Adding first-vs-second-half controls to the training data raises specificity from 30.2 to 83.7 and changes F1 from 39.2 to 37.1.
Adding all per-object questions of \ours as auxiliary tasks gives 40.5 F1.

\paragraph{Two-view training data.}
Without the two-frame questions and the KC-3D data, the verifier reaches 59.9 AP and 61.3 F1 on SD-V, 50.6 F1 on SD-K and 63.7 AP50 on RC-3D.
With both sources it reaches 57.6, 61.1, 52.7 and 64.3, so a single model covers multi-view and two-view input.

\section{Result analysis}
\label{sec:analysis}

\begin{figure}[t]
\centering
\includegraphics[width=0.7\linewidth]{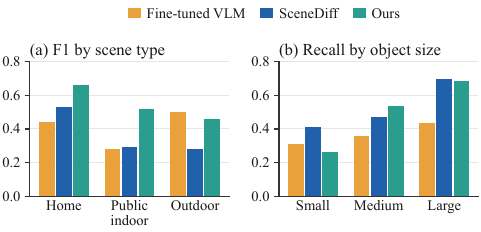}
\caption{\textbf{Breakdown} on the SD-V test split. (a) F1 by scene type, with the largest gain in cluttered public indoor scenes such as stores, offices and gyms. (b) Recall by object size, where small objects (29 of 1,016) remain hard.}
\label{fig:breakdown}
\end{figure}

\begin{figure}[t]
\centering
\includegraphics[width=0.7\linewidth]{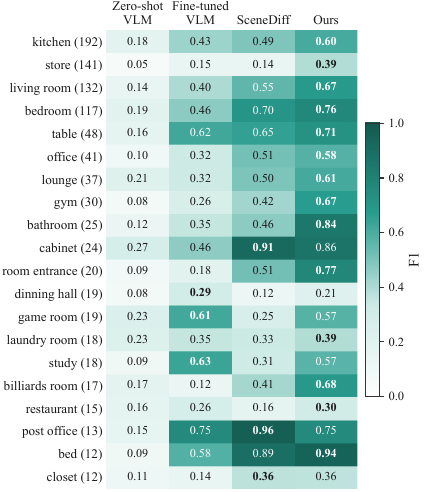}
\caption{\textbf{F1 per scene category} on the SD-V test split for each method (number of changed objects in brackets, best per row in bold).}
\label{fig:cats}
\end{figure}

\paragraph{Change type, object size and scene type.}
\cref{fig:breakdown} breaks the results down by scene type and object size.
\ours outperforms SceneDiff in home, public indoor and outdoor scenes, with the largest gain in public indoor scenes, where F1 rises from 29 to 52.
Recall is 54\% on medium and 69\% on large objects.
Small objects, whose largest visible area is below $64\times64$ pixels, are only 29 in the test split and reach a recall of 26\%, so higher-resolution views of small objects are a direct way to improve.
\cref{fig:cats} further reports F1 for 20 scene categories.
\ours is best in 14 of them and raises stores from 14 F1 with SceneDiff to 39.

\begin{figure}[t]
\centering
\includegraphics[width=0.7\linewidth]{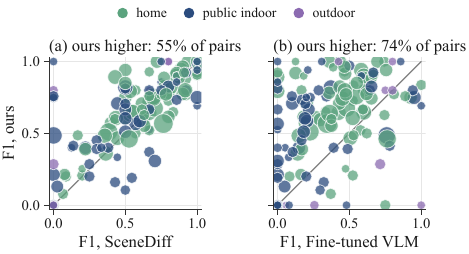}
\caption{\textbf{Per-pair F1} of \ours against SceneDiff (a) and the fine-tuned VLM (b) on the SD-V test split. Marker size: number of changed objects. Colour: scene type. Points above the diagonal are pairs where \ours is better.}
\label{fig:scenes}
\end{figure}

\paragraph{Per-scene comparison.}
\cref{fig:scenes} plots the F1 of \ours against SceneDiff and against holistic fine-tuning for every test video pair.
\ours is better than SceneDiff on 55\% of the pairs and better than holistic fine-tuning on 74\%, with the gains concentrated in public indoor scenes.

\begin{figure}[t]
\centering
\includegraphics[width=0.7\linewidth]{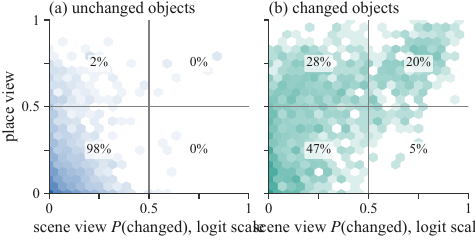}
\caption{\textbf{The two views complement each other.} Scene-view against place-view probability of changed for SceneDiff candidates on the SD-V test split, for unchanged (a) and changed (b) objects, with the share of candidates in each quadrant at 0.5.}
\label{fig:views}
\end{figure}

\paragraph{Complementarity of the two views.}
\cref{fig:views} shows the probabilities that the two views assign to SceneDiff candidates.
Of the unchanged candidates, 97.8\% are judged unchanged by both views.
Of the changed candidates, 20\% receive a high probability from both views, 28\% from the place view only and 5\% from the scene view only.
The place view recovers many changes that the scene view cannot confirm, and the scene view supplies the context of whether the place was captured at all.

\begin{figure}[t]
\centering
\includegraphics[width=0.7\linewidth]{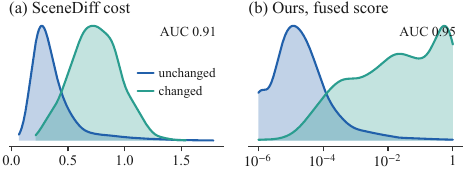}
\caption{\textbf{Score distributions} of unchanged and changed SceneDiff candidates under the SceneDiff cost (a) and the fused score of \ours (b), with the AUC of separating the two.}
\label{fig:dist}
\end{figure}

\begin{figure}[t]
\centering
\includegraphics[width=0.7\linewidth]{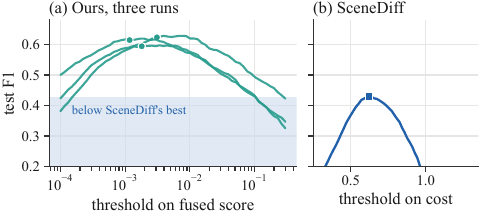}
\caption{\textbf{Threshold sensitivity} on the SD-V test split for \ours (a, three runs) and SceneDiff (b). Dots: thresholds chosen on dev. Shaded: F1 below the best value of SceneDiff.}
\label{fig:thr}
\end{figure}

\paragraph{Score distributions and threshold sensitivity.}
\cref{fig:dist} compares the distributions of unchanged and changed candidates under the SceneDiff cost and the fused score of \ours, which separates them with an AUC of 0.95 instead of 0.91.
\cref{fig:thr} plots test F1 against the threshold.
Over three orders of magnitude, from $10^{-4}$ to $10^{-1}$, all three runs of \ours stay above 43, the best F1 that SceneDiff reaches at any threshold, and the thresholds chosen on dev lie on the plateau of the curve.

\section{Full record of learned registration}
\label{sec:learnedfull}

\begin{table*}[t]
\centering
\caption{\textbf{Learned registration} (one run each). The VLM predicts the place itself as a frame and a box in the other capture, optionally refined on a zoomed crop.
Places found: agreement with the $\pi^3$ place in the frame the model chose (IoU$\geq$0.5), for changed / unchanged objects.
$\Delta$: system AP with the learned place view minus that with the $\pi^3$ place view under the same verifier.
The main system with $\pi^3$ places reaches 57.6 / 46.9 AP. $^\dagger$SFT on half of the training scenes and RL on the other half, compared with its own $\pi^3$ reference of 51.4 / 43.2 AP.}
\label{tab:learned}
\footnotesize
\begin{tabular}{@{}lccccc@{}}
\toprule
Version & places found (\%) & SD-V AP & SD-K AP & $\Delta$ SD-V & $\Delta$ SD-K \\
\midrule
locate in the 10-frame layout & 6.8 / 32.2 & 56.4 & 41.9 & -3.5 & -5.1 \\
+ zoomed refinement & 12.0 / 48.5 & 54.5 & 41.9 & -2.9 & -4.9 \\
object masked (context only) & 13.1 / 33.2 & 54.9 & 39.4 & -3.4 & -6.9 \\
+ changed $\times$4, KC-3D localisation & 19.6 / 50.9 & 53.6 & 39.9 & -3.4 & -4.2 \\
+ wide context, second pass & 22.5 / 50.6 & 53.8 & 42.6 & -1.3 & -4.5 \\
+ RL, verification reward (v2b) & 21.7 / 51.1 & 53.0 & 42.6 & -2.1 & -4.5 \\
cross-fitted RL on unseen real scenes$^\dagger$ & 14.0 / 41.6 & 49.2 & 37.7 & -2.2 & -5.5 \\
\bottomrule
\end{tabular}
\end{table*}

\begin{figure}[t]
\centering
\includegraphics[width=0.7\linewidth]{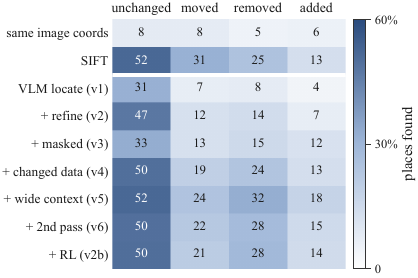}
\caption{\textbf{Learned registration by change type.} Share of places found (\%) for unchanged, moved, removed and added objects, for the unmapped box, SIFT matching and each learned version.}
\label{fig:learnedtype}
\end{figure}

\cref{tab:learned} and \cref{fig:learnedtype} detail all learned registration variants of \cref{sec:learned}.
The variants build on two stages.
The first stage takes a hypothesis in the scene view and outputs a frame index and a box in the other capture.
The second stage crops the region around the first-stage box, shows it next to an object crop with wider context, and outputs a refined box.
All variants are supervised by the $\pi^3$ places of the training videos.
\cref{fig:learnedtype} breaks the places found down by change type.
Added objects are the hardest, at 4\% to 18\% across variants, removed objects reach at most 32\%, and unchanged objects lie between 31\% and 52\%.

\begin{figure*}[t]
\centering
\includegraphics[width=\linewidth]{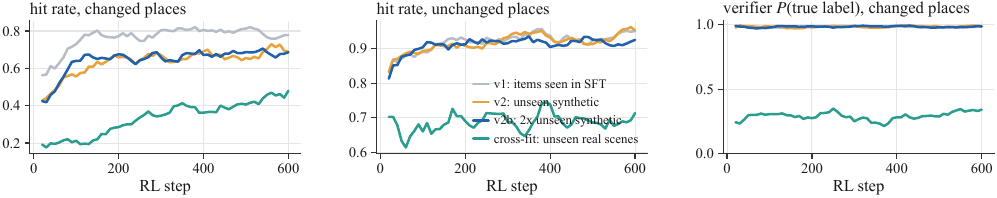}
\caption{\textbf{RL for learned localisation} (50-step moving average) in four settings: items seen in SFT (v1), unseen synthetic items (v2), twice as many unseen synthetic items (v2b), and cross-fitting on unseen real scenes. With seen or synthetic items the verifier reward saturates near 1, while on unseen real scenes it stays informative and the hit rate on changed places rises during training. No setting improves the test split (\cref{tab:learned}).}
\label{fig:rl}
\end{figure*}

Reinforcement learning uses the second stage as the policy, with LoRA of rank 32, 8 samples per question with group-normalised advantages, learning rate $10^{-5}$ and 600 steps.
The reward is
\begin{multline}
r=\mathrm{IoU}(\hat{b},b^{\pi^3})+0.5\cdot\mathbb{1}[\mathrm{IoU}\ge0.5]\\
+\lambda\cdot\mathbb{1}[\mathrm{IoU}\ge0.3]\cdot q_{\theta_0}\big(y\mid V_{\text{place}}(\hat{b})\big),
\end{multline}
where the last term is the probability that the frozen verifier gives to the correct answer at the predicted place, and the IoU gate keeps the policy from collecting the verification reward at wrong places.
\cref{fig:rl} shows the training curves of four RL settings.
On items seen in SFT and on synthetic data, the verification reward is close to saturation.
Under cross-fitting, where the policy is fine-tuned on one half of the training videos and trained with RL on the other, unseen half, the verification reward on changed objects lies between 0.1 and 0.4 and gives a useful learning signal.
The policy raises the hit rate on changed objects of these training scenes from 27\% to 63\%, while the hit rate on the test split is 14.0\%, on par with 14.8\% before RL, and a control run with the IoU reward alone gives the same result.

\section{Inference cost}
\label{sec:cost}

\cref{tab:cost} lists the runtime per SD-V test item by stage, measured on one A100 with model loading included.
SceneDiff itself takes about 25 seconds, most of it for SAM segmentation.
$\pi^3$ registers the places of all candidates in under 1 second, and the verifier needs about 2.7 seconds for the two views of about 11 candidates, an increase of about 11\% in total.

\begin{table}[t]
\centering
\caption{\textbf{Inference cost} per SD-V test item (5+5 frames, about 11 proposals) on one A100, model loading included.}
\label{tab:cost}
\footnotesize
\begin{adjustbox}{max width=\linewidth}
\begin{tabular}{@{}lc@{}}
\toprule
Stage & seconds / item \\
\midrule
SceneDiff ($\pi^3$ + SAM + DINOv3; shared with the baseline) & 25 \\
$\pi^3$ places for all proposals & $<$1 \\
verifier, scene view + place view (Qwen3.5-9B) & 2.7 \\
\midrule
total, ours & $\approx$28 (+11\%) \\
\bottomrule
\end{tabular}
\end{adjustbox}
\end{table}

\section{Qualitative results}
\label{sec:morequal}

\begin{figure*}[t]
\centering
\includegraphics[width=\linewidth]{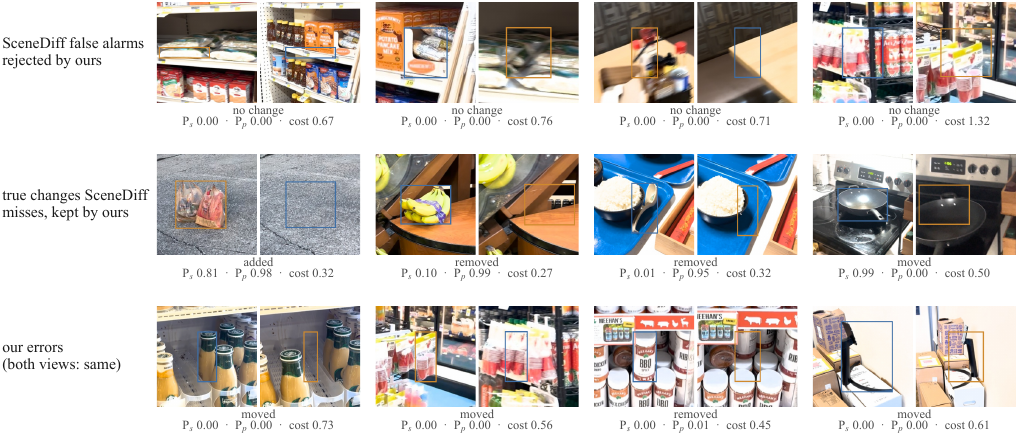}
\caption{\textbf{Place-view cases} on the SD-V test split (four random cases per row). Each case shows the object crop and the place crop with the scene-view and place-view probabilities of changed ($P_s$, $P_p$) and the SceneDiff cost.
Top: SceneDiff false alarms rejected by \ours, mostly on dense shelves and motion-blurred frames.
Middle: true changes that SceneDiff ranks low and \ours keeps.
Bottom: errors of \ours, three of four on store shelves where an identical product stands at the registered place.}
\label{fig:placeview}
\end{figure*}

\cref{fig:placeview} shows typical place-view cases, with four candidates drawn at random per row.
The first row shows candidates to which SceneDiff gives a high cost and which \ours judges unchanged.
They mostly come from densely stocked shelves and motion-blurred frames, and the two crops show the same content.
The second row shows true changes to which SceneDiff gives a low cost and which \ours keeps, including bags newly placed on the ground, removed bananas and a removed spoon, and a moved pan.
The third row shows errors of \ours, three of them on store shelves.
When one product is taken from a shelf, identical products still stand right at its place, so the place view looks unchanged.
\cref{fig:responses} lists the raw outputs of the zero-shot VLM, the VLM fine-tuned on the same data and \ours on the scene of \cref{fig:teaser}, scored by the matching rule of the benchmark.

\begin{figure*}[t]
\centering
\footnotesize
\begin{tabular}{@{}p{0.31\linewidth}p{0.31\linewidth}p{0.31\linewidth}@{}}
\toprule
\textbf{Zero-shot Qwen3-VL-8B} & \textbf{Qwen3.5-9B fine-tuned on the same data} & \textbf{\ours{}} \\
\midrule
\xmark~added: mirror \textcolor{black!50}{(v2 f1)} & \xmark~removed: yellow bottle \textcolor{black!50}{(v1 f2)} & \cmark~added: gray chair \textcolor{black!50}{(v2 f2, score 0.140)} \\
\xmark~added: light stand \textcolor{black!50}{(v2 f1)} & \xmark~removed: tissue \textcolor{black!50}{(v1 f3)} & \cmark~added: white chair \textcolor{black!50}{(v2 f2, score 0.062)} \\
\xmark~added: laptop \textcolor{black!50}{(v2 f3)} & \xmark~removed: cardboard box \textcolor{black!50}{(v1 f4)} & \cmark~added: laptop \textcolor{black!50}{(v2 f5, score 0.046)} \\
\xmark~added: power strip \textcolor{black!50}{(v2 f4)} & \xmark~removed: toolbox \textcolor{black!50}{(v1 f4)} & \dmark~added: white chair \textcolor{black!50}{(v2 f1, score 0.026)} \\
\xmark~removed: vertical blinds \textcolor{black!50}{(v1 f1)} & \xmark~moved: white chair \textcolor{black!50}{(v1 f5)} & \dmark~added: laptop \textcolor{black!50}{(v2 f3, score 0.013)} \\
\xmark~removed: plant \textcolor{black!50}{(v1 f1)} & \xmark~removed: curtain \textcolor{black!50}{(v1 f1)} & \dmark~added: gray chair \textcolor{black!50}{(v2 f1, score 0.009)} \\
\xmark~moved: plant \textcolor{black!50}{(v2 f3)} & \xmark~added: mirror \textcolor{black!50}{(v2 f1)} &  \\
\midrule
\textcolor{black!60}{0 correct, 7 false} & \textcolor{black!60}{0 correct, 7 false} & \textcolor{black!60}{3 correct, 0 false, 3 repeated} \\
\bottomrule
\end{tabular}
\caption{\textbf{Raw outputs on the scene of \cref{fig:teaser}} (SD-V test split, 5+5 frames). Ground truth: four moved objects (white chair, gray chair, laptop, floor lamp).
\cmark/\xmark{} follow the benchmark rule (named frame, box IoU$\geq$0.3 with a changed object, each object counted once), and \dmark{} marks a repeated hit on an object already found, which is not counted.
Holistic VLMs name objects that only one capture shows (curtain, mirror, light stand) and miss the moved chairs. The outputs of \ours{} are the SceneDiff hypotheses kept by the verifier, each named by the object it matches.}
\label{fig:responses}
\end{figure*}

\section{Relation to dense change detection benchmarks}
\label{sec:dense}

Benchmarks such as PASLCD~\cite{galappaththige2025multi} and that of 3DGS-CD~\cite{lu20253dgs} address a different setting.
Their first capture covers the scene from all directions, the viewpoints of the second capture mostly lie within the range of the first, and evaluation measures pixel-level change masks.
In this setting, methods that optimise 3D Gaussians per scene and compare renderings perform best.
SceneDiff addresses two independent, sparse captures along different trajectories and evaluates object-level changes.
This is the situation of a robot that revisits a space, and it is the target setting of \ours.

\section{Limitations and outlook}
\label{sec:limits}

The three stages of \ours are handled by a proposer, a 3D reconstruction and a VLM, and each can be replaced directly by a stronger model.
The changes that can be found are bounded by the candidates of the proposer.
SceneDiff covers about three quarters of the changed objects on SD-V, switching from a generic detector to SceneDiff already raises AP from 43.0 to 57.4, and proposers with higher recall will bring further gains.
The accuracy of the place depends on the 3D reconstruction.
In textureless or very dark frames, its uncertainty propagates to the place box, and the fast progress of feed-forward reconstruction models will directly improve this stage.
Verification runs on crops of fixed resolution, in which very small objects cover only a few pixels, and zooming the crop adaptively with the object size or using VLMs with higher input resolution are natural extensions.
Finally, \ours reuses the association of the same object across the two captures from the proposer.
Letting the VLM verify a moved object at its old and new place together promises a further gain in classifying the change type of moved objects.

\end{document}